\documentclass{article}

\usepackage{PRIMEarxiv}

\usepackage[utf8]{inputenc} % allow utf-8 input
\usepackage[T1]{fontenc}    % use 8-bit T1 fonts
\usepackage{hyperref}       % hyperlinks
\usepackage{url}            % simple URL typesetting
\usepackage{booktabs}       % professional-quality tables
\usepackage{amsfonts}       % blackboard math symbols
\usepackage{nicefrac}       % compact symbols for 1/2, etc.
\usepackage{microtype}      % microtypography
\usepackage{lipsum}
\usepackage{fancyhdr}       % header
\usepackage{graphicx}       % graphics
\graphicspath{{media/}}     % organize your images and other figures under media/ folder

\usepackage{booktabs}
\usepackage{dingbat}
\usepackage{multirow}
\usepackage{mathtools}

\usepackage{pifont}% http://ctan.org/pkg/pifont
\newcommand{\cmark}{\ding{51}}%
\newcommand{\xmark}{\ding{55}}%

\usepackage{caption} 
\title{Foveation in the Era of Deep Learning
\thanks{\textit{Accepted to The 34\textsuperscript{th} British Machine Vision Conference (BMVC2023)}
}}

\author{
  George Killick, Paul Henderson, Paul Siebert, Gerardo Aragon-Camarasa \\
  School of Computer Science \\
  University of Glasgow \\
  \texttt{2182951k}@student.gla.ac.uk} 
\begin{document}
\maketitle

\begin{abstract}
In this paper, we tackle the challenge of actively attending to visual scenes using a foveated sensor. We introduce an end-to-end differentiable foveated active vision architecture that leverages a graph convolutional network to process foveated images, and a simple yet effective formulation for foveated image sampling. Our model learns to iteratively attend to regions of the image relevant for classification. We conduct detailed experiments on a variety of image datasets, comparing the performance of our method with previous approaches to foveated vision while measuring how the impact of different choices, such as the degree of foveation, and the number of fixations the network performs, affect object recognition performance. We find that our model outperforms a state-of-the-art CNN and foveated vision architectures of comparable parameters and a given pixel or computation budget. The source code is publicly available at \url{https://github.com/georgeKillick90/FovConvNeXt}.
\end{abstract}

\section{Introduction}

 Many biological vision systems sense the world with a foveated sensor, where the highest resolution processing is limited to only a small central portion of the visual field (the fovea). Computer vision systems have taken inspiration from this aspect of biological vision and incorporated it into visual attention models that learn to sample and process visual scenes actively \cite{mnih2014recurrent, ba2014multiple, sermanet2014attention}. The promise of foveated vision is the ability to resolve and process fine details while simultaneously maintaining a wide field of view, which has applications to problems where semantic information can exist over a high-dynamic range of scales. More generally, it is well known that scaling the resolution of inputs to CNNs can reliably improve accuracy in objection recognition problems \cite{tan2019efficientnet}.
Through sparse sampling in the periphery of the field of view, foveated sensors can achieve this with significantly fewer pixels than a uniform sensor, making it an appealing approach to building parsimonious vision systems. This comes with the caveat that a visual attention mechanism must be incorporated into the system, in order to guide high-resolution processing to areas of interest.

In practice, it has proven difficult to reconcile foveated sensors with modern computer vision systems. Due to their space-variant resolution, foveated sensors are not naturally amenable to standard convolution operators which expect uniformly sampled inputs. Many works propose sensors that can be conveniently mapped back to grids to circumvent this problem \cite{lukanov2021biologically, esteves2018polar} but in doing so, inherently change the equivariance properties of the network. There is evidence, such as in the case of the log-polar mapping, that these properties may be sub-optimal for object recognition \cite{ozimek2019space, torabian2020comparison, lukanov2021biologically}. Other works propose approximations of foveated sensors by rescaling multiple crops of differing fields of view (FoV) \cite{mnih2014recurrent, harris2019foveated}. These methods are again convenient, but raise questions on how features from different crops should interact at intermediate stages of processing. Other challenges arise from training an attention mechanism to guide the sensor. Reinforcement learning has been the method of choice for several works \cite{mnih2014recurrent, ba2014multiple, elsayed2019saccader}, however this is accompanied by additional challenges due to the difficulty of training systems with this method from scratch.

To this end, we propose a foveated convolutional neural network architecture that actively attends to scenes to perform image classification. Our work addresses the aforementioned challenges in the following ways. We introduce a graph convolutional approach for processing foveated images, removing the need for any mapping to a uniform sampling arrangement, in turn affording more flexible control over the equivariance properties of the layer. Additionally, this facilitates the use of a single foveated feature map, allowing foveal and peripheral features to interact at intermediate stages of the network. We adopt a differentiable sampling mechanism based on \cite{jaderberg2015spatial} allowing end-to-end training of the entire system, avoiding reinforcement learning. Additionally, we propose a novel foveated sensor that can be easily configured to control the sampling resolution across the visual field. We verify the efficacy of our model through several experiments. Crucially, we show that our model can be applied effectively to object recognition tasks on challenging real-world image datasets. In summary, we make the following main contributions:
\begin{itemize}
    \item A novel end-to-end differentiable foveated vision architecture which is able to outperform previous foveated architectures by atleast 1\% and up to 3.5\% on the Imagenet100 dataset, as well as a state-of-the-art CNN at an equivalent number of input pixels and FLOPs (Section \ref{sec:arch}).

    \item A novel graph convolution layer designed to process foveated images without requiring them to be mapped to a uniform grid (Section \ref{sec:graphconv})

    \item We show that foveated vision architectures are much better at recognizing objects over a wide range of scales than uniform non-attentive vision architectures when constrained to the same number of input pixels (Section \ref{sec:experiments}).
\end{itemize}

\section{Related Work}

Visual hard attention models employ a dynamic sensor and learn an attention policy to guide the sensor to informative regions of a scene for downstream tasks. Mnih et al.'s seminal work \cite{mnih2014recurrent} uses reinforcement learning to train a recurrent neural network that guides a foveated sensor (approximated by crops of increasing FoV) to classify MNIST digits in cluttered backgrounds. Ba et al. \cite{ba2014multiple} and Sermanet et al. \cite{sermanet2014attention} extend this work to multiple object recognition and fine-grained classification of natural images respectively. Li et al \cite{li2017dynamic} introduce a dynamic stopping condition to a visual attention model, allowing it to terminate computation early if the model is significantly confident in its prediction, in turn reducing computational costs. This work foregoes a foveated sensor in favour of an adaptive zoomable sampling grid. Elsayed et al. \cite{elsayed2019saccader} learn an attention policy over the output of a BagNet \cite{brendel2019approximating} and propose a pretraining routine that can alleviate the difficulty of training visual attention models with reinforcement learning. In a similar approach Rangrej et al \cite{rangrej2022consistency} train a transformer to sequentially select its input patches. Unlike \cite{elsayed2019saccader}, their approach does not require the full image to be processed by the network. They additionally introduce a student-teacher training paradigm, where the teacher model has access to the whole image. These works have been primarily concerned with how to train hard-attention models and their applications to different visual perception tasks but provide relatively few insights into the design of the foveated sensor and feature extractor. We instead, provide greater focus on the underlying feature extraction process and leverage a comparitively simpler end-to-end differentiable approach.

Differentiable image sampling techniques \cite{jaderberg2015spatial, gregor2015draw} offer a method to train visual attention models without reinforcement learning. Spatial Transformer Networks (STNs) \cite{jaderberg2015spatial, angles2021mist} train a neural network to predict affine transformations of the input image to a classifier. Recasens et al \cite{recasens2018learning} and Thavamani et al \cite{thavamani2021fovea} propose similar extensions to STNs for object recognition and detection respectively. They use adaptive sampling grids that sample an input image at higher resolution based on a saliency map computed by a CNN. Foveated variants of spatial transformers have been proposed. Harris et al \cite{harris2019foveated} use multiple crops, similar to \cite{mnih2014recurrent}, to improve object localization. Polar Transformer Networks \cite{esteves2018polar} use the log-polar mapping to achieve rotation and scale equivariance. Cheung et al \cite{cheung2016emergence} train a recurrent neural network to attend to and classify MNIST digits. They introduce a learnable sampling lattice and show the emergence of a foveated structure.
Many works avoid explicitly learning an attention mechanism for a particular task. For instance,\cite{karpathy2014large, li2017foveanet} uses a multi-fov crop to reduce the number of input pixels to a video classification CNN. Ozimek et al \cite{ozimek2019space} use a biologically plausible foveated sensor based on \cite{schwartz1980computational} to perform image classification, exploiting camera bias in the dataset to circumvent the need for an attention mechanism. Lukanov et al \cite{lukanov2021biologically} adopt the foveated sensor proposed in \cite{martinez2006new} and compute consecutive fixations through argmax over class activation maps. Jonnalagadda et al \cite{jonnalagadda2023foveater} apply foveated pooling to a small ResNet \cite{he2016deep} feature extractor before passing the pooled representations to a vision transformer; subsequent fixations are determined by the maximal activation in the final self-attention layer. 

These works encompass a broad range of foveated and adaptive downsampling techniques however relatively few comparisons between different methods has been made. Additionally, many methods report results on simple datasets such as MNIST which may not represent their efficacy on harder natural image datasets. We conduct comparisons across a variety of different approaches on the Imagenet100 dataset. Furthermore, we identify possible aspects of previous approaches to incorporating foveated vision into CNNs, specifically with regard to equivariance properties and the interaction between foveal and peripheral features, that may inhibit their performance. We address this with a novel graph convolution layer for foveated images and show improvements in object recognition performance as a result.

\section{Method}

\begin{figure}
    \centering
    \includegraphics[width=11cm]{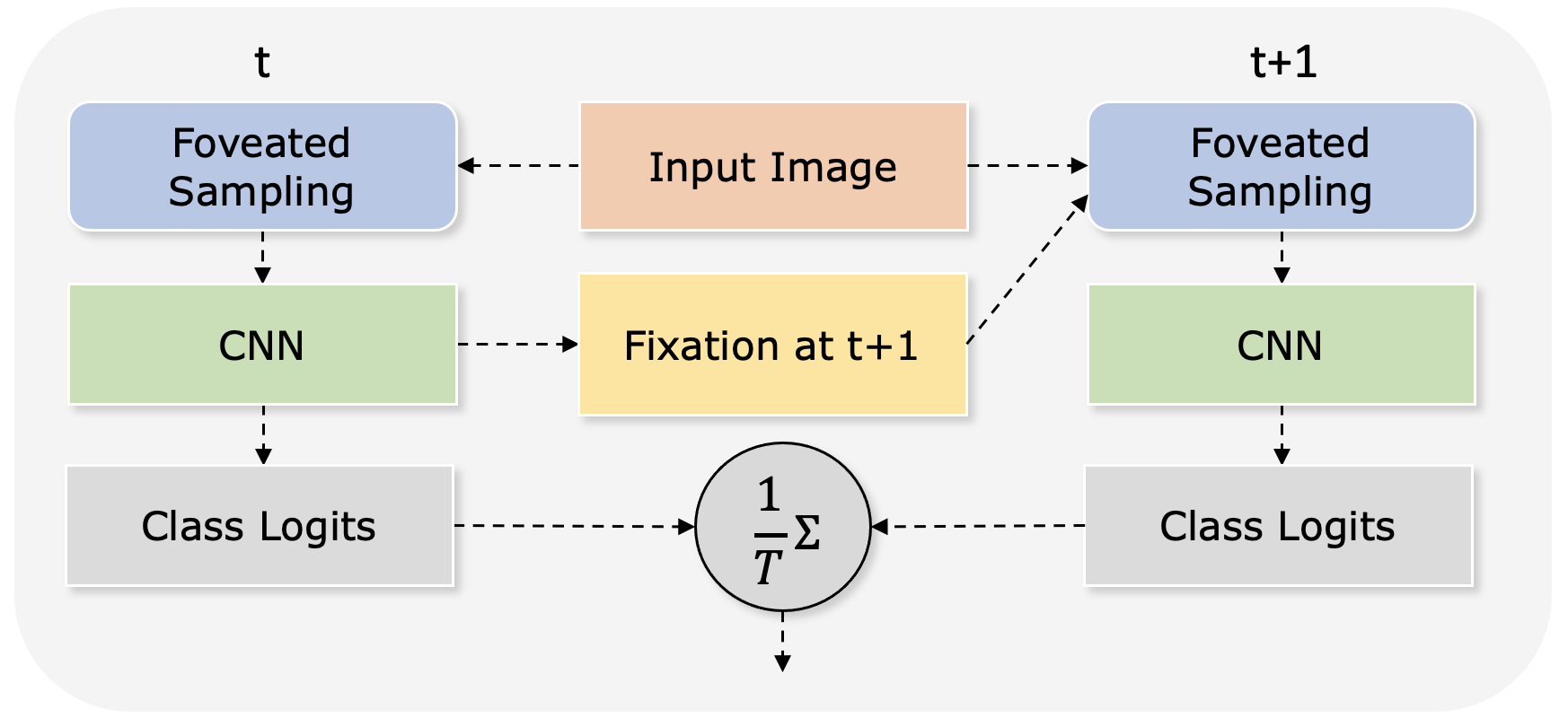}
    \caption{Our pipeline for sequential foveated image classification. The input image is sampled by our foveated sensor and processed by a novel graph convolutional network, which predicts class logits, and optionally a fixation location center the sensor on in the next time step.}
    \label{fig:architecture}
\end{figure}

\subsection{Architecture\label{sec:arch}}

We introduce a foveated graph convolutional architecture that learns to attend to salient areas of images to perform object recognition. We consider two variants of this architecture, a sequential model (Figure \ref{fig:architecture}) that can repeatedly attend to the image for many time-steps, and a spatial transformer variant which uses a separate localization network to provide a single location (fixation) for the foveated classifier to attend to.

These architectures share in common four core components, a differentiable foveated sensor, a graph convolutional feature extractor, an attention module and a classifier head. The foveated sensor samples a uniform input image centred on a given fixation coordinate. This foveated image is subsequently processed by a graph convolutional feature extractor to produce a multi-channel feature map. A conventional classifier head consisting of a global average pooling layer and linear layer produces a class prediction. The attention module is similar to that proposed in Polar Transformer Networks \cite{esteves2018polar}. It takes as input a \textit{d}-dimensional featuremap and applies a $1 \times 1$ convolution to produce a single channel saliency map. We apply softmax activation to the saliency map and multiply each element by its corresponding (x,y) coordinate. Finally we sum these values to arrive at a final fixation coordinate. This is equivalent to finding the expected coordinate using the probability of a given location containing salient information.

In the sequential variant, at time step 0, the network receives an initial fixation at the center of the image. A class prediction is computed via the method described above. The feature maps computed by the graph convolutional network are fed into the attention module to compute a fixation for the next time step. This allows the network to leverage rich features already computed by the feature-extractor for classification to inform the next fixation with minimal overhead. The number of fixations the network is allowed to perform is given by the hyperparameter $T$. After predictions for each time step have been computed, a final prediction is made by averaging predictions over all time steps. In the spatial transformer variant, an initial fixation is computed by applying our attention module to the output of a separate localization network that operates on a uniformly downsampled input image. This fixation is then fed into the foveated classifier and produces a class prediction. Both methods are trained end-to-end via gradient descent through the backpropagation of a standard cross-entropy classification loss.

\subsection{Foveated Sensor\label{sec:foveated}}
\begin{figure}
    \centering
    \includegraphics[width=15cm]{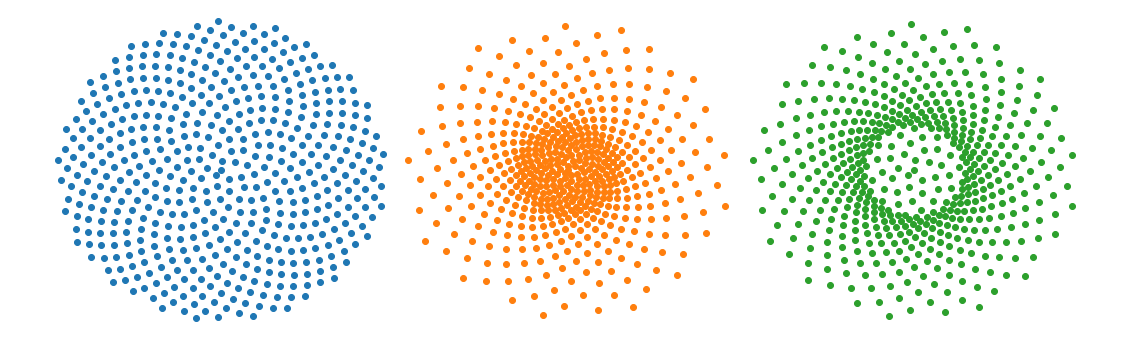}
    \vspace{-12pt}
    \caption{Left to Right: Vogel's model of a sunflower capitulum \cite{vogel79}, our foveated adaptation (eq.~\ref{eq:sunflower}) where fovea sampling density is well parameterised, and our adaptation where the fovea sampling density is poorly parameterised.}
    \label{fig:foveated_sensor}
\end{figure}

We define a foveated arrangement of bilinear sampling kernels over the image plane to perform foveated sampling. We adopt Vogel's model of seed arrangement in a sunflower capitulum \cite{vogel79} to provide an easy way to pack points approximately uniformly in a circle (Fig. \ref{fig:foveated_sensor}). Vogel's model is simple and computed at initialisation; it also has the property that the distribution of pixels in a small patch (i.e. the size of a convolutional filter) is approximately isotropic. We adapt this model to a foveated arrangement of points with the highest sampling density in the centre and smoothly decaying resolution as eccentricity increases. We achieve this by logarithmically spacing points outside a given radius of the fovea. Formally the position of the $i^{th}$ sampling kernel in polar coordinates $(\rho, \theta)$ is given by:
\begin{equation}
\label{eq:sunflower}
    \theta_{i} = 2 \pi i \phi, \;\;\;\;\;\;
    \lambda = r^\frac{1}{d-N}, \;\;\;\;\;\;
    \rho_{i} = \begin{cases}
        r\sqrt{\frac{i}{d}}, & \text{if $i<d - 1$}\\
        r\lambda^{i-d}, & \text{otherwise}.
    \end{cases}
  % \rho_{i} =\begin{cases}
  %   %r\sqrt{\frac{i}{d}}, & \text{if $i<d -1 $}.\\
  %   %{r\lambda^{i-d}}, & \text{otherwise}.
  % \end{cases}
\end{equation}
where $\phi$ is the golden ratio, $N$ is the total number of sampling kernels, $d$ is the number of sampling kernels in the fovea, and $r$ is the radius of the fovea. The sampling density of the fovea can be controlled independently from the size of the fovea; however, setting the number of sampling kernels in the fovea too low will result in an excessively sparse fovea. This is undesirable as the sampling resolution will not be able to resolve details in the fovea. (Figure~\ref{fig:foveated_sensor}). We explore different values for $r$ in Section \ref{sec:ablation}.%that the sampling resolution in the fovea is approximately equal to the sampling density immediately outside the fovea (see Section \ref{sec:ablation}). 

In order to translate the sensor over the input image, we apply an $(x, y)$ offset to all sampling kernels. Backpropagation with respect to this offset is then possible by accumulating the gradients of each individual sampling kernel output with respect to its position \cite{jaderberg2015spatial, esteves2018polar}. For cases where sampling kernels extend beyond the image boundary, we use border padding to fill in missing pixels. We found that zero padding could cause the network to collapse to predict the same fixation for all images (Supplementary 3).

\paragraph{Alternative Foveated Sensors.}
We consider two alternative methods for producing foveated images in our experiments. Firstly, a sensor comprised of multiple crops of increasing field of view, sampled at the resolution of the smallest crop \cite{mnih2014recurrent, ba2014multiple, harris2019foveated}. For models processing this type of image, we process each crop independently and average class predictions over crops at the end. We experimented with stacking each crop in the channel dimension before being processed by the CNN but found the former to work better. We also test the log-polar transform as a foveation method \cite{esteves2018polar, traver2010review}. In this case, convolutions use circular padding in the angular dimension of the image to address the discontinuity introduced in the log-polar transform \cite{esteves2018polar, kim2020cycnn}.

\subsection{Graph Convolutions\label{sec:graphconv}}

\begin{figure}
    \centering
    \includegraphics[width=15cm]{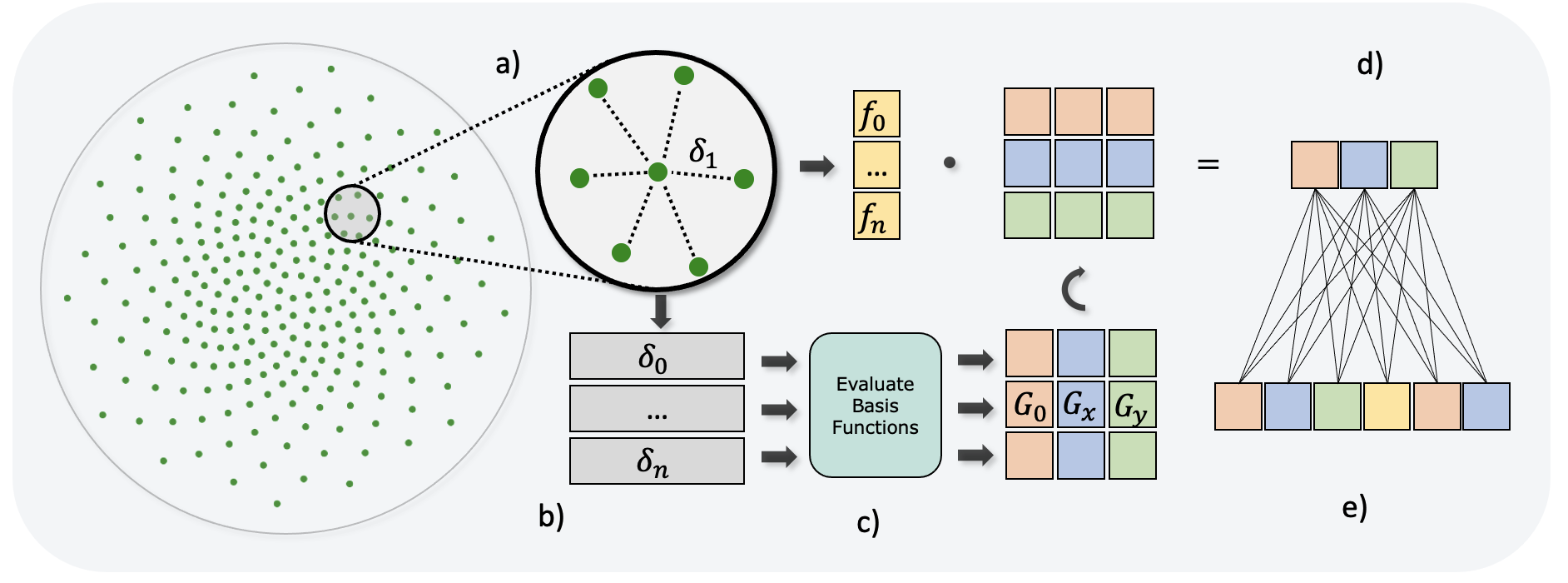}
    \caption{Our novel graph CNN for processing foveated images. Pixels are sampled non-uniformly (a) and for each, its nearest neighbours are found and their relative displacements $\delta_i$ are calculated (b). Gaussian Derivative basis functions are evaluated using the $\delta_i$ offsets to obtain basis filter weights for the current patch (c). A weighted sum between basis filter weights and features is computed to obtain a vector of responses to the basis functions (d). A learned linear combination of basis function responses computes final feature vectors for a given patch (e).}
    \label{fig:convoperator}
\end{figure}

The arrangement of foveated sampling kernels does not follow a regular grid and, as such, it poses a challenge for downstream processing since typical 2D convolution operators cannot be applied here. Unlike standard images, local patches of pixels on a foveated image may not have a canonical ordering or spatial arrangement of points. This necessitates a convolution operator that can produce filter responses that are largerly invariant to changes in these properties. We overcome this by leveraging a graph convolution operator (Fig. \ref{fig:convoperator}) that is invariant to the permutation of pixels in the receptive field, and computes filter weights based on their relative position to the receptive field's center.

\paragraph{Graph Construction.} Our convolution layer takes as input a set of feature vectors, each associated with a vertex $u \in U$ of a 2D planar graph and a position $(x, y)$ that defines the feature's spatial position within the feature map.
Similarly, we define a set of vertices $V$ pertaining to output features with an $(x,y)$ coordinate and a feature vector $\hat{f}$ that we aim to compute.
For each output vertex $v_{i} \in V$, we define its receptive field $R_{i}$ as its $K$ nearest spatial neighbours in $U$ and connect them to form a bipartite graph $G = (U, V, E)$.
We label each edge in the graph with the spatial offset between the two vertices it connects. For example, the edge $e_{ij}$ connecting an input vertex $u_{i} \in U$ to output vertex $v_{j} \in V$ is labelled as $\delta_{ij}=(x_{i}-x_{j}, y_{i}-y_{j})$. We additionally normalize the $\delta$ offsets by the mean $\delta$ offset for a given patch. This scales filters as sampling density becomes more sparse in the periphery. Scaling filters exponentially with eccentricity produces feature maps that are equivariant to global scale transformations of the input. This is seen in applying convolutions to log-polar images \cite{esteves2018polar}. However, our method differs in that our convolution maintains a consistent filter orientation with respect to the cartesian axes.

\paragraph{Edge Conditioned Filter Weights.}

In order to allow our model to learn filters that are as expressive as ordinary convolutions, our convolution layer produces filter weights as a function of edge labels ($\delta$ offsets). This allows the relative spatial positions of features to be considered, permitting the learning of filters such as oriented edge detectors. Graph convolution operators of this form, often termed edge-conditioned graph convolutions \cite{simonovsky2017dynamic}, use a parameterised differentiable function approximator to map edge labels to filter weights \cite{wu2019pointconv, boulch2020convpoint, hermosilla2018monte}. We instead decompose the filter weights into a parameterised linear combination of Gaussian derivative basis functions (Figure \ref{fig:basis} in supplementary material), which has been shown to work well for learning convolution filters \cite{jacobsen2016structured, lindeberg2022scale} uniform images. This has two major benefits for our approach. Firstly, we can explicitly control the maximum frequency of our filters by changing the $\sigma$ of the Gaussian derivatives or truncating the higher-order derivatives. Secondly, the basis is known a priori and fixed, meaning we can precompute all basis filters at network initialisation. The convolution operation for a single-channel input and single-channel output is defined in the supplementary material (Section 2).

\section{Experiments\label{sec:experiments}}

We conduct our main experiments on ImageNet-100, a 100-class subset of the ImageNet-1K dataset \cite{imagenet}, to test the efficacy of our method against alternative and similar methods in the literature (Section \ref{sec:exp-effective}). We conduct additional experiments on an augmented MNIST dataset, to evaluate our model's ability to classify very large and very small objects under a highly constrained number of pixels (Section \ref{sec:mnist}). Finally, we perform ablation studies (Section \ref{sec:ablation}) on our model using the Imagewoof dataset \cite{imagewoof}, a 10-class fine-grained classification subset of ImageNet-1K comprising different breeds of dogs.

\subsection{Effectiveness of foveation on image classification tasks\label{sec:exp-effective}}

We perform experiments on the ImageNet-100 dataset, comprised of 130,000 images across 100 categories. We constrain the number of input pixels to the feature extractors to be $112^2$ for all models. The reduced sampling resolution requires the models to adapt their sampling grid to salient regions of the input image that would ordinarily be lost through uniform downsampling. Additionally, the feature extractors must be able to extract rich features from the images in order to perform the classification task effectively.

\paragraph{Baselines.} 

For all baseline methods, we use the atto variant of ConvNeXt as our feature extraction backbone \cite{woo2023convnext, liu2022convnet}. 
Models that receive an input of fewer than $224^{2}$ pixels have their final downsampling layer in the convolution stages removed to maintain approximately $7^2$ pixels in the final feature maps. We consider three non-attentive baselines, the default ConvNeXt at $224\times224$ input resolution, a downsampled input variant at $112\times112$ input resolution, and a foveated variant using our sensor and graph convolution layers. We consider 6 variants of Spatial Transformer Networks (STN). For the localisation networks, we use a truncated ConvNeXt atto, which has all layers after the penultimate convolution stage removed. All localization networks operate on a downsampled $112\times112$ input image. For the full affine spatial transformer, we append a 2 hidden layer MLP, with 128 neurons in each layer, batch normalization and ReLU activation, that regresses the affine transformation matrix. For the foveated variants, we consider the Polar Transformer Network \cite{esteves2018polar} (PTN), a multi-fov crop STN \cite{harris2019foveated}, an STN with the FCG sensor \cite{martinez2006new} and our foveated sensor with our graph convolution operator. In these variants, we append our attention module (Section \ref{sec:arch}) to compute fixations. Finally, we compare against the Learning to Zoom model. This method remains largely unchanged from its original proposal except from updating the localization network and feature extractor from ResNet \cite{he2016deep} to ConvNeXt \cite{woo2023convnext}.

\begin{table}
\resizebox{1\textwidth}{!}{\begin{tabular}{lllccccc}
\hline
\textbf{Method} & \textbf{Operator} & \textbf{Sensor} & \textbf{\# Input Pixels} & \textbf{\# Fixations} & \textbf{Params (M)} & \textbf{GFLOPs} & \textbf{Accuracy (\%)} \\ \hline
ConvNeXt \cite{liu2022convnet, woo2023convnext} & Conv & Uniform & 50176 & - & 3.7 & 0.55 & 78.4 \\
ConvNeXt & Conv & Uniform & 12544 & - & 3.7 & 0.20 & 70.0 \\
Ours (non-attentive) & Graph Conv & Our Sensor & 12544 & - & 3.7 & 0.20 & 72.5 \\ \hline
STN \cite{jaderberg2015spatial} & Conv & Uniform & 12544 & 1 & 4.8 & 0.32 & 72.7 \\
PTN \cite{esteves2018polar} & Conv & Log-Polar & 12800 & 1 & 4.8 & 0.33 & 70.7 \\
FCG-STN \cite{martinez2006new} & Conv & FCG  & 12544 & 1 & 4.8 & 0.33 & 71.0 \\
Fov STN \cite{harris2019foveated} & Conv & Multi-FoV Crops & 12800 & 1 & 4.8 & 0.33 & 71.8 \\
Fov STN (ours) & Graph Conv & Our Sensor & 12544 & 1 & 4.8 & 0.32 & 74.2 \\
Learning to Zoom \cite{recasens2018learning} & Conv & Deformable Grid & 12544 & 1 & 4.8 & 0.32 & 75.8 \\ \hline
Sequential & Conv & FCG \cite{martinez2006new} & 12800 & 2 & 3.7 & 0.41 & 70.2 \\
Sequential & Conv & Log-Polar & 12800 & 2 & 3.7 & 0.41 & 70.4 \\
Sequential & Conv & Multi-FoV Crops & 12800 & 2 & 3.7 & 0.41 & 72.8 \\
Sequential (Ours) & Graph Conv & Our Sensor & 12544 & 2 & 3.7 & 0.41 & 73.8 \\
Sequential (Ours) & Graph Conv & Our Sensor & 12544 & 3 & 3.7 & 0.61 & 76.5 \\ \hline
\end{tabular}}
\caption{Top-1 Accuracy on the Imagenet100 test set. We split the table into three sections. Top: non-attentive models. Middle: Spatial Transformer like models. Bottom: Sequential Models.}
\label{tab:imagenet}
\end{table}

\paragraph{Results and discussion.}

We report top-1 accuracy on the Imagenet-100 test set in Table \ref{tab:imagenet}, along with the number of parameters and GFLOPs. Even without attention, we find that our foveated graph ConvNeXt outperforms a uniform ConvNeXt by 2.5\%. We posit that this performance improvement is possible due to the tendency of relevant objects to be centred in the frame. Notably, despite not having attention, this model also outperforms all alternative foveated spatial transformer methods by atleast 0.7\% and up to 1.8\% in the case of the log-polar sensor. When incorporating attention into our method, further increase over other foveated architectures is observed. Our sequential and spatial transformer variants perform atleast 1.0\% and 2.4\% better than their next best foveated counterparts. In both cases, the next best method is the Multi-FoV crop method which shares similarities with our method in that filters are only translated and scaled over the visual field, however, unlike our method, foveal and peripheral features are maintained in separate featuremaps. We find that performance of our model trails slightly behind the learning to zoom method and 'full' resolution ConvNeXt. By allowing the sequential model to perform more fixations we can achieve higher accuracy than Learning-to-Zoom (76.5\% vs 75.8\%) albeit at an increased number of FLOPs. Similarly, given the performance improvement of foveated sampling over uniform downsampling for non-attentive models, we speculate that a similar increase in accuracy could be obtained for a foveated sensor with $224^2$ pixels.

\subsection{Performance with Diverse Object Scales\label{sec:mnist}}

A key benefit of foveated sampling is the ability to resolve fine details yet also reason about a wide field of view, with fewer pixels than uniform sampling.
We analyse this claim by performing experiments using two augmented variants of the MNIST dataset, where we closely control the scale distribution of digits and avoid any bias regarding the positions of objects. This property is not common in image classification datasets (e.g. Imagenet \cite{imagenet}), which typically exhibit camera bias -- a tendency for objects of interest to be centred in the frame and relatively large in proportion to the full image.
Our 2 MNIST variants, namely S-MNIST and ST-MNIST (scaled and scaled-translated, respectively), consist of randomly scaled MNIST digits between 1 and 8 times their original size. In S-MNIST, we place the scaled digits at the centre of a $224\times224$ pixel canvas, whereas in ST-MNIST, we also randomly translate the digit. Crucially, while the images for this task are $224\times224$ in size, we enforce a constraint that all networks aggressively downsample the input to approximately $28^{2}$ pixels to significantly reduce the computational overhead of the system. In such a scenario, small-scaled digits will have a large portion of salient information about their class destroyed, and attentive methods must learn to adapt their sensors to resolve and classify small digits. Full details of network architectures and implementation details are given in the supplementary material.

\paragraph{Results and discussion.}

Our results show that adaptive sampling methods can accurately solve the classification task and significantly improve over the uniform downsampling approach (Table \ref{tab:mnist}). In particular, our active approach achieves an 11\% performance improvement over the uniform downsampling baseline on ST-MNIST and does so with only 66 additional parameters. We show that the adaptive methods can nearly match their ST-MNIST performance showing they accurately learn to locate and attend to small objects in the scene. For additional context, we evaluate our approach while only fixating on the image centre. This model achieves only 79.2\% accuracy, showing that attention is indeed needed to solve this task. The full affine spatial transformer performs worse than other adaptive methods by $\sim7\%$ on ST-MNIST and $\sim5\%$ on S-MNIST. We observed that training would frequently collapse for spatial transformers, we include results for runs that did not collapse. Notably, the full affine spatial transformer is the only method that adapts its sampling grid by regressing transformation parameters through an MLP. This suggests that a principled computation of transformation parameters performed in the other models is significantly easier to optimise. All the adaptive methods are approximately on par with each other.

\begin{table}
\resizebox{1\textwidth}{!}{\begin{tabular}{llccc}
\hline
\textbf{Method} & \textbf{Sensor} & \textbf{Separate Localisation Network} & \multicolumn{2}{c}{\textbf{Accuracy (\%)}} \\ \hline
 &  &  & \textbf{S-MNIST} & \textbf{ST-MNIST} \\ \hline
CNN & Uniform & \xmark & 86.6 & 86.9 \\
STN \cite{jaderberg2015spatial} & Uniform & \cmark & 93.7 & 90.1 \\
PTN \cite{esteves2018polar} & Log-Polar & \cmark & 98.4 & 97.5 \\
Fov-STN \cite{harris2019foveated} & Multi-FoV Crops & \cmark & 98.2 & 96.9 \\
Fov-STN (Ours) & Our Sensor & \cmark & 98.4 & 97.6 \\
Sequential (Ours) & Our Sensor & \xmark & 98.4 & 97.9 \\
Graph-CNN (Ours & Our Sensor & \xmark & 98.3 & 79.2 \\ \hline
\end{tabular}}
\caption{Accuracy on our MNIST variants with high-scale variation. We show that adaptive and foveated sampling methods significantly improve the models ability to recognize objects over a wide range of scales compared to static uniform methods (CNN)}
\label{tab:mnist}
\end{table}

\subsection{Ablation Experiments\label{sec:ablation}}

We conduct additional experiments on Imagewoof, a 10-class subset of Imagenet, to probe various aspects of our model. This dataset contains 10 different species of dogs with approximately 900 images per class in training and 400 per class for testing. We take 10\% of the training set to create a validation set which we used to tune hyperparameters and select the best checkpoint. The fine-grained nature of this dataset means it remains challenging, while its small size allows us to rapidly perform experiments in a variety of settings.
We examine three different aspects of our model -- how the radius of the fovea, the number of fixations and different policies for attention affect classification accuracy. We maintain the same experimental set-up and training routine as the Imagenet-100 experiments, except we change the learning rate from 0.004 to 0.0025.

\paragraph{Effect of Fovea Radius.} In order to ascertain whether performance increase over a uniform baseline is due to foveation, we sweep over a range of fovea radii. All models are trained with three fixations. We consider 3 variants of our model for this experiment, a $112^{2}$ pixel input variant, a $56^{2}$ pixel variant, and a narrow $56^{2}$ pixel variant with half the number of filters in each layer. We find optimal fovea radius values of 60\%, except in the case of the narrow model which is 40\%. The benefit of foveation is more pronounced for the scaled down models. Sandler et al \cite{sandler2019non} observe that, due to the common architectural design choices of CNNs, resolution and width scaling are approximately the same. This could explain why foveation becomes a more important factor when scaling the model down with respect to these dimensions and suggests foveation can provide an increasingly beneficial role in more lightweight CNN architectures.

\begin{figure}
    \centering
    \includegraphics[width=0.49\linewidth]{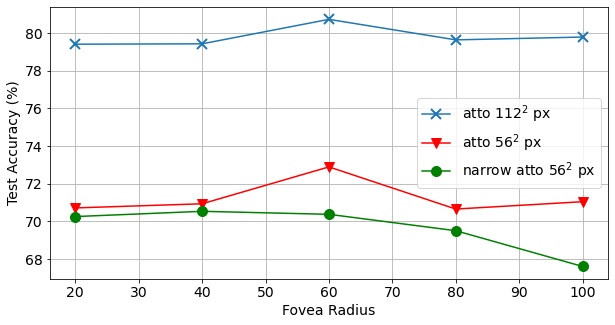}\hfill
    \includegraphics[width=0.49\linewidth]{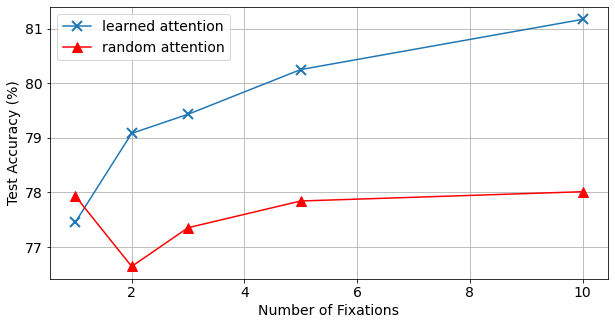}
    \caption{\textbf{Left:} Performance of our model with varying radius of fovea (100 corresponds to negligible foveation). \textbf{Right:} Performance with varying numbers of fixations; we see that performance increases with more fixations when the attention policy is learned}
    \label{fig:graphs}
\end{figure}

\paragraph{Effect of Number of Fixations.} We investigate how our model scales as the number of fixations increases (Figure \ref{fig:graphs}, right). We consider two different attention policies, our learned attention, and random attention. We show that under a learned attention policy our network continues to result in improved classification accuracy as the number of fixations increases. Noticeably there is an approximately 4\% accuracy boost from 1 to 10 fixations. By contrast, the network does not derive significant benefits from extra fixations under a random policy, and even degrades when the number of fixations is low. Given the intuition that object recognition improves when the foveated sensor is centered on the object of interest, we should expect it to decrease when it is not. This can explain the drop in performance with a random policy as our model simply averages predictions across timesteps. 

\section{Conclusion\label{sec:conclusion}}

In this work, we have presented an end-to-end differentiable active vision model. Our model actively attends to visual scenes and aggregates information over multiple timesteps through a space-variant foveated sensor. We show that our model can outperform a non-attentive CNN at classifying natural images by $\sim2\%$ with 1 fixation using a $112^2$ pixel input and performs better than alternative foveated CNNs. 
We further show that foveated architectures have particular application to recognizing objects over a high dynamic range of scales. 
Ablation experiments show that  foveation becomes increasingly beneficial for smaller architectures suggesting they can fulfil a useful role in lightweight computer vision systems. There are many avenues for future work with regard to this architecture. More powerful methods for integrating information from multiple fixations could prove beneficial (e.g. self-attention). Additionally, allowing the network to remember previously attended locations and prevent returning to them may help in maximizing the information the system gains when fixating over many time steps.

%Bibliography
\bibliographystyle{unsrt}  
\bibliography{references}  

\begin{thebibliography}{10}

\bibitem{mnih2014recurrent}
Volodymyr Mnih, Nicolas Heess, Alex Graves, and Koray Kavukcuoglu.
\newblock Recurrent models of visual attention.
\newblock In Zoubin Ghahramani, Max Welling, Corinna Cortes, Neil~D. Lawrence, and Kilian~Q. Weinberger, editors, {\em Advances in Neural Information Processing Systems 27: Annual Conference on Neural Information Processing Systems 2014, December 8-13 2014, Montreal, Quebec, Canada}, pages 2204--2212, 2014.

\bibitem{ba2014multiple}
Jimmy Ba, Volodymyr Mnih, and Koray Kavukcuoglu.
\newblock Multiple object recognition with visual attention.
\newblock In Yoshua Bengio and Yann LeCun, editors, {\em 3rd International Conference on Learning Representations, {ICLR} 2015, San Diego, CA, USA, May 7-9, 2015, Conference Track Proceedings}, 2015.

\bibitem{sermanet2014attention}
Pierre Sermanet, Andrea Frome, and Esteban Real.
\newblock Attention for fine-grained categorization.
\newblock In Yoshua Bengio and Yann LeCun, editors, {\em 3rd International Conference on Learning Representations, {ICLR} 2015, San Diego, CA, USA, May 7-9, 2015, Workshop Track Proceedings}, 2015.

\bibitem{tan2019efficientnet}
Mingxing Tan and Quoc Le.
\newblock Efficientnet: Rethinking model scaling for convolutional neural networks.
\newblock In {\em International conference on machine learning}, pages 6105--6114. PMLR, 2019.

\bibitem{lukanov2021biologically}
Hristofor Lukanov, Peter K{\"o}nig, and Gordon Pipa.
\newblock Biologically inspired deep learning model for efficient foveal-peripheral vision.
\newblock {\em Frontiers in Computational Neuroscience}, 15:746204, 2021.

\bibitem{esteves2018polar}
Carlos Esteves, Christine Allen{-}Blanchette, Xiaowei Zhou, and Kostas Daniilidis.
\newblock Polar transformer networks.
\newblock In {\em 6th International Conference on Learning Representations, {ICLR} 2018, Vancouver, BC, Canada, April 30 - May 3, 2018, Conference Track Proceedings}. OpenReview.net, 2018.

\bibitem{ozimek2019space}
Piotr Ozimek, Nina Hristozova, Lorinc Balog, and Jan~Paul Siebert.
\newblock A space-variant visual pathway model for data efficient deep learning.
\newblock {\em Frontiers in cellular neuroscience}, 13:36, 2019.

\bibitem{torabian2020comparison}
Parsa Torabian, Ronak Pradeep, Jeff Orchard, and Bryan Tripp.
\newblock Comparison of foveated downsampling techniques in image recognition.
\newblock {\em Journal of Computational Vision and Imaging Systems}, 6(1):1--3, 2020.

\bibitem{harris2019foveated}
Ethan William~Albert Harris, Mahesan Niranjan, and Jonathon Hare.
\newblock Foveated convolutions: improving spatial transformer networks by modelling the retina.
\newblock In {\em Shared Visual Representations in Human and Machine Intelligence: 2019 NeurIPS Workshop}, 2019.

\bibitem{elsayed2019saccader}
Gamaleldin Elsayed, Simon Kornblith, and Quoc~V Le.
\newblock Saccader: Improving accuracy of hard attention models for vision.
\newblock {\em Advances in Neural Information Processing Systems}, 32, 2019.

\bibitem{jaderberg2015spatial}
Max Jaderberg, Karen Simonyan, Andrew Zisserman, et~al.
\newblock Spatial transformer networks.
\newblock {\em Advances in neural information processing systems}, 28, 2015.

\bibitem{li2017dynamic}
Zhichao Li, Yi~Yang, Xiao Liu, Feng Zhou, Shilei Wen, and Wei Xu.
\newblock Dynamic computational time for visual attention.
\newblock In {\em Proceedings of the IEEE International Conference on Computer Vision Workshops}, pages 1199--1209, 2017.

\bibitem{brendel2019approximating}
Wieland Brendel and Matthias Bethge.
\newblock Approximating cnns with bag-of-local-features models works surprisingly well on imagenet.
\newblock In {\em 7th International Conference on Learning Representations, {ICLR} 2019, New Orleans, LA, USA, May 6-9, 2019}. OpenReview.net, 2019.

\bibitem{rangrej2022consistency}
Samrudhdhi~B Rangrej, Chetan~L Srinidhi, and James~J Clark.
\newblock Consistency driven sequential transformers attention model for partially observable scenes.
\newblock In {\em Proceedings of the IEEE/CVF Conference on Computer Vision and Pattern Recognition}, pages 2518--2527, 2022.

\bibitem{gregor2015draw}
Karol Gregor, Ivo Danihelka, Alex Graves, Danilo Rezende, and Daan Wierstra.
\newblock Draw: A recurrent neural network for image generation.
\newblock In {\em International conference on machine learning}, pages 1462--1471. PMLR, 2015.

\bibitem{angles2021mist}
Baptiste Angles, Yuhe Jin, Simon Kornblith, Andrea Tagliasacchi, and Kwang~Moo Yi.
\newblock Mist: Multiple instance spatial transformer.
\newblock In {\em Proceedings of the IEEE/CVF Conference on Computer Vision and Pattern Recognition}, pages 2412--2422, 2021.

\bibitem{recasens2018learning}
Adria Recasens, Petr Kellnhofer, Simon Stent, Wojciech Matusik, and Antonio Torralba.
\newblock Learning to zoom: a saliency-based sampling layer for neural networks.
\newblock In {\em Proceedings of the European Conference on Computer Vision (ECCV)}, pages 51--66, 2018.

\bibitem{thavamani2021fovea}
Chittesh Thavamani, Mengtian Li, Nicolas Cebron, and Deva Ramanan.
\newblock Fovea: Foveated image magnification for autonomous navigation.
\newblock In {\em Proceedings of the IEEE/CVF international conference on computer vision}, pages 15539--15548, 2021.

\bibitem{cheung2016emergence}
Brian Cheung, Eric Weiss, and Bruno~A. Olshausen.
\newblock Emergence of foveal image sampling from learning to attend in visual scenes.
\newblock In {\em 5th International Conference on Learning Representations, {ICLR} 2017, Toulon, France, April 24-26, 2017, Conference Track Proceedings}. OpenReview.net, 2017.

\bibitem{karpathy2014large}
Andrej Karpathy, George Toderici, Sanketh Shetty, Thomas Leung, Rahul Sukthankar, and Li~Fei-Fei.
\newblock Large-scale video classification with convolutional neural networks.
\newblock In {\em Proceedings of the IEEE conference on Computer Vision and Pattern Recognition}, pages 1725--1732, 2014.

\bibitem{li2017foveanet}
Xin Li, Zequn Jie, Wei Wang, Changsong Liu, Jimei Yang, Xiaohui Shen, Zhe Lin, Qiang Chen, Shuicheng Yan, and Jiashi Feng.
\newblock Foveanet: Perspective-aware urban scene parsing.
\newblock In {\em Proceedings of the IEEE International Conference on Computer Vision}, pages 784--792, 2017.

\bibitem{schwartz1980computational}
Eric~L Schwartz.
\newblock Computational anatomy and functional architecture of striate cortex: a spatial mapping approach to perceptual coding.
\newblock {\em Vision research}, 20(8):645--669, 1980.

\bibitem{martinez2006new}
Jos{\'e} Mart{\'\i}nez and Leopoldo~Altamirano Robles.
\newblock A new foveal cartesian geometry approach used for object tracking.
\newblock {\em SPPRA}, 6:133--139, 2006.

\bibitem{jonnalagadda2023foveater}
Aditya Jonnalagadda, William~Yang Wang, B.S. Manjunath, and Miguel Eckstein.
\newblock Foveater: Foveated transformer for image classification, 2023.

\bibitem{he2016deep}
Kaiming He, Xiangyu Zhang, Shaoqing Ren, and Jian Sun.
\newblock Deep residual learning for image recognition.
\newblock In {\em Proceedings of the IEEE conference on computer vision and pattern recognition}, pages 770--778, 2016.

\bibitem{vogel79}
Helmut Vogel.
\newblock A better way to construct the sunflower head.
\newblock {\em Bellman Prize in Mathematical Biosciences}, 44:179--189, 1979.

\bibitem{traver2010review}
V~Javier Traver and Alexandre Bernardino.
\newblock A review of log-polar imaging for visual perception in robotics.
\newblock {\em Robotics and Autonomous Systems}, 58(4):378--398, 2010.

\bibitem{kim2020cycnn}
Jinpyo Kim, Wooekun Jung, Hyungmo Kim, and Jaejin Lee.
\newblock Cycnn: A rotation invariant cnn using polar mapping and cylindrical convolution layers.
\newblock {\em arXiv preprint arXiv:2007.10588}, 2020.

\bibitem{simonovsky2017dynamic}
Martin Simonovsky and Nikos Komodakis.
\newblock Dynamic edge-conditioned filters in convolutional neural networks on graphs.
\newblock In {\em Proceedings of the IEEE conference on computer vision and pattern recognition}, pages 3693--3702, 2017.

\bibitem{wu2019pointconv}
Wenxuan Wu, Zhongang Qi, and Li~Fuxin.
\newblock Pointconv: Deep convolutional networks on 3d point clouds.
\newblock In {\em Proceedings of the IEEE/CVF Conference on computer vision and pattern recognition}, pages 9621--9630, 2019.

\bibitem{boulch2020convpoint}
Alexandre Boulch.
\newblock Convpoint: Continuous convolutions for point cloud processing.
\newblock {\em Computers \& Graphics}, 88:24--34, 2020.

\bibitem{hermosilla2018monte}
Pedro Hermosilla, Tobias Ritschel, Pere-Pau V{\'a}zquez, {\`A}lvar Vinacua, and Timo Ropinski.
\newblock Monte carlo convolution for learning on non-uniformly sampled point clouds.
\newblock {\em ACM Transactions on Graphics (TOG)}, 37(6):1--12, 2018.

\bibitem{jacobsen2016structured}
Jorn-Henrik Jacobsen, Jan Van~Gemert, Zhongyu Lou, and Arnold~WM Smeulders.
\newblock Structured receptive fields in cnns.
\newblock In {\em Proceedings of the IEEE Conference on Computer Vision and Pattern Recognition}, pages 2610--2619, 2016.

\bibitem{lindeberg2022scale}
Tony Lindeberg.
\newblock Scale-covariant and scale-invariant gaussian derivative networks.
\newblock {\em Journal of Mathematical Imaging and Vision}, 64(3):223--242, 2022.

\bibitem{imagenet}
Jia Deng, Wei Dong, Richard Socher, Li-Jia Li, Kai Li, and Li~Fei-Fei.
\newblock {ImageNet}: A large-scale hierarchical image database.
\newblock In {\em IEEE Conference on Computer Vision and Pattern Recognition (CVPR)}, pages 248--255, 2009.

\bibitem{imagewoof}
fastai.
\newblock Imagenette and imagewoof.

\bibitem{woo2023convnext}
Sanghyun Woo, Shoubhik Debnath, Ronghang Hu, Xinlei Chen, Zhuang Liu, In~So Kweon, and Saining Xie.
\newblock Convnext {V2:} co-designing and scaling convnets with masked autoencoders.
\newblock In {\em {IEEE/CVF} Conference on Computer Vision and Pattern Recognition, {CVPR} 2023, Vancouver, BC, Canada, June 17-24, 2023}, pages 16133--16142. {IEEE}, 2023.

\bibitem{liu2022convnet}
Zhuang Liu, Hanzi Mao, Chao{-}Yuan Wu, Christoph Feichtenhofer, Trevor Darrell, and Saining Xie.
\newblock A convnet for the 2020s.
\newblock In {\em {IEEE/CVF} Conference on Computer Vision and Pattern Recognition, {CVPR} 2022, New Orleans, LA, USA, June 18-24, 2022}, pages 11966--11976. {IEEE}, 2022.

\bibitem{sandler2019non}
Mark Sandler, Jonathan Baccash, Andrey Zhmoginov, and Andrew Howard.
\newblock Non-discriminative data or weak model? on the relative importance of data and model resolution.
\newblock In {\em Proceedings of the IEEE/CVF International Conference on Computer Vision Workshops}, pages 0--0, 2019.

\bibitem{loshchilov2017decoupled}
Ilya Loshchilov and Frank Hutter.
\newblock Decoupled weight decay regularization.
\newblock In {\em 7th International Conference on Learning Representations, {ICLR} 2019, New Orleans, LA, USA, May 6-9, 2019}. OpenReview.net, 2019.

\bibitem{loshchilov2016sgdr}
Ilya Loshchilov and Frank Hutter.
\newblock {SGDR:} stochastic gradient descent with warm restarts.
\newblock In {\em 5th International Conference on Learning Representations, {ICLR} 2017, Toulon, France, April 24-26, 2017, Conference Track Proceedings}. OpenReview.net, 2017.

\bibitem{muller2021trivialaugment}
Samuel~G M{\"u}ller and Frank Hutter.
\newblock Trivialaugment: Tuning-free yet state-of-the-art data augmentation.
\newblock In {\em Proceedings of the IEEE/CVF international conference on computer vision}, pages 774--782, 2021.

\end{thebibliography}

\appendix

\section{Appendix}

\subsection{Gaussian Derivative Basis Filters}

The basis filters can be computed through the multiplication of a Gaussian windowing function $G$ and the Hermite polynomials $H$ where $H_{m}(x)$ computes the $m^{th}$ order partial derivative along the $x$ axis. Accordingly, a 2-D Gaussian Derivative basis filter can be computed as follows:

\begin{equation} \label{eq:basis}
    B(x, y, \sigma, m) = (-1)^{m_x m_y} H_{m_x}\left(\frac{x}{\sigma \sqrt{2}}\right) H_{m_y} \left(\frac{y}{\sigma \sqrt{2}}\right) G(x, y, \sigma)
\end{equation}

Where $x$ and $y$ are the coordinates at which the filter is evaluated, $\sigma$ controls the size of the filter, and $m = (m_x, m_y)$ defines the order of the partial derivatives in the $x$ and $y$ directions respectively. The Hermite polynomials are defined as: 

\begin{equation}
\begin{aligned}
    & H_0(x) = 1 \\
    & H_1(x) = 2x \\
    & H_2(x) = 4x^2 - 2\\
    & H_3(x) = 8x^3 - 12x \\
    & H_4(x) = 16x^4 - 48x^2 + 12 \\
    & H_5(x) = 32x^5 - 160x^3 + 120x 
\end{aligned}
\end{equation}

\begin{figure}[t]
    \centering
    \includegraphics[width=6cm]{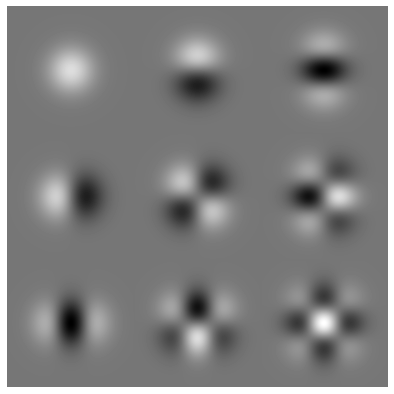}
    \caption{Visualization of Gaussian Derivative Basis Functions. The order of the partial derivative in the $y$ direction increases from left to right, and the $x$ direction increases from top to bottom. Our choice of basis functions for a given layer is determined by the hyperparameter $M$. The basis includes all Gaussian derivatives where the maximum order of their partial derivatives is $\le M$}
    \label{fig:basis}
\end{figure}

\subsection{Gaussian Derivative Graph Convolution}

Equation \ref{eq:conv} defines our convolution operation for single channel input and single channel output layer. 

\begin{equation} \label{eq:conv}
    \hat{f}_{j} = \sum_{m=0}^{|S|} w_{m} \sum_{i \in K_j}^{} f_{i} B(\delta_{xij}, \delta_{yij}, \sigma_{j}, S_m)
\end{equation}

$f_i$ is the feature associated with the vertex $u_i \in U$ in the input features, and $w$ is a set of learned weights that weigh the responses to individual basis filters. $B$ refers to equation \ref{eq:basis} that computes basis filter weights. $\delta_{xij}$ and $\delta_{yij}$ are the delta offset labels associated with the edge $e_{ij} \in E$. $\sigma_j$ is the size of the filter when centred on the output node $v_j \in V$. $S = \{(0,0), (0, 1) ... (M,M)\}$ where $S_m$ defines the order of the partial derivatives in the $x$ and $y$ directions, for the $m^{th}$ basis filter. $M$ is a hyperparameter that defines the maximum order of a partial derivative that is permitted for the basis filters, i.e. any filter with a partial derivative greater than $M$ is not included in the basis. $K_j$ refers to the neighbourhood of the output vertex $v_j \in V$, i.e. the set of vertices in $U$ that have an edge with $v_j$. Additionally, we normalise the filter values for each basis filter by subtracting its mean and dividing by its $\ell_2$ norm for a given neighbourhood $K$. One exception is for the $0^{th}$ order gaussian derivative, i.e. a normal gaussian filter, which we normalize by dividing by its $\ell_1$ norm. 

\subsection{Border vs. Zero Padding}

In early pilot testing, we observed that zero padding would frequently lead to the collapse of the attention module (i.e. it would predict the same fixation regardless of the input image). We conduct an experiment comparing how this affects performance using a 2 fixation sequential network on the Imagewoof. We observe a decrease from 79.2\% accuracy to 77.8\%. This is approximately in line with the performance improvement from a 1 fixation network to a 2 fixation network (Figure. \ref{fig:graphs}). We could not ascertain the exact cause of this peculiarity. Fortunately, border padding is readily provided in most frameworks meaning this is not problematic.

\subsection{Training Details (ImageNet100)}
Networks are trained with a batch size of 64, and the AdamW \cite{loshchilov2017decoupled} optimizer. We perform a linear warmup on the learning rate for 5 epochs, followed by cosine annealing for 85 epochs \cite{loshchilov2016sgdr}. During training, we use trivial data augmentation \cite{muller2021trivialaugment}, followed by resizing the shortest side to 256px and a random resized crop of $224\times224$. At test time, we resize the shortest side to 256px and perform a $224\times224$ centre crop. Images are normalized using Imagenet mean and standard deviation. We use a learning rate of 0.004 and a weight decay of 0.005 for the feature extractors. For the localization network, we use a learning rate of 0.0004 for the final $1\times1$ convolution (or MLP in case of the full affine spatial transformer) and a learning rate of 0.00004 for the convolution stages as done in \cite{recasens2018learning}. We found many methods that use a localisation network collapse during training; therefore, we use weights pre-trained on Imagenet-1K for the localisation networks. We evaluate the model on the test set using the best-performing model checkpoint on the validation set. We include the implementation details that are specific to our graph convolutional ConvNeXt atto in table \ref{table:config}.

% \usepackage{booktabs}
\begin{table}
\centering
\begin{tabular}{@{}l|c@{}}
\toprule
\begin{tabular}[c]{@{}l@{}}Foveated Graph\\ ConvNeXt atto config\end{tabular} & \begin{tabular}[c]{@{}c@{}}Imagenet 100k\\ Settings\end{tabular} \\ \midrule
Fovea Radius                                                                  & 40\%                                                             \\
Stem - kernel size                                                            & 16                                                               \\
Stem - sigma                                                                  & 1.0                                                              \\
Stem - max order                                                              & 4                                                                \\
Blocks - kernel size                                                          & 49                                                               \\
Blocks - sigma                                                                & 0.8                                                              \\
Blocks - max order                                                            & 4                                                                \\
Downsampling - kernel size                                                    & 4                                                                \\
Downsampling - sigma                                                          & 0.6                                                              \\
Downsampling - max order                                                      & 0                                                                \\ \bottomrule
\end{tabular}

\caption{Stem refers to the initial convolution layer in the ConvNeXt architecture, Blocks refers to the configuration of the depthwise convolution layers in the NeXt Blocks. Downsampling refers to the configuration of the downsampling layers that reduce spatial dimensionality between stages in the ConvNeXt architecture. Kernel size is analogous to kernel size in ordinary convolution layers. Sigma determines the size of Gaussian derivative basis filters; max order refers to the maximum order of partial derivatives used in the basis.}
\label{table:config}
\end{table}

\subsection{Implementation details (MNIST)}
We utilise a 4-layer feature extractor for all methods, with 64 $3\times3$ filters in each layer, followed by batch normalisation and ReLU activation. We found that larger networks did not increase performance. For our method, we use our graph convolution in place of the standard 2D convolution operator and a filter size of 9 so that the filters are the same size as in the standard grid convolution case. The convolutional features from the final layer are global average pooled and passed to a linear classifier. We restrict the number of input pixels to the feature extractor to be $28^{2}$, the size of a normal MNIST image. We use the same architecture as the feature extractor, without average pooling and the linear classifier, for methods that use a separate localisation network to adjust the sampling grid. The localisation networks receive a $28\times28$ uniformly down-sampled image as input.

We independently perform a random hyperparameter search across learning rate and weight decay for all methods. We use the AdamW optimiser \cite{loshchilov2017decoupled} and train with a batch size of 64 for 20 epochs, which we found was sufficient for networks to converge. We use a 5-epoch linear warm-up schedule followed by 15-epoch cosine decay \cite{loshchilov2016sgdr}. We report accuracy on the test set using the best-performing model checkpoint on the validation set.

\end{document}